\documentclass[10pt,twocolumn,letterpaper]{article}

\usepackage{cvpr}
\usepackage{times}
\usepackage{epsfig}
\usepackage{graphicx}
\usepackage{amsmath}
\usepackage{amssymb}
\usepackage{subfig}

\usepackage{kotex}

\usepackage[pagebackref=true,breaklinks=true,letterpaper=true,colorlinks,bookmarks=false]{hyperref}

\cvprfinalcopy 

\def\cvprPaperID{****} 

\ifcvprfinal\fi
\begin{document}

\definecolor{mypink1}{rgb}{0.858, 0.188, 0.478}
\title{Exploring Unlabeled Faces for Novel Attribute Discovery}

\author{
Hyojin Bahng$^1$
~~Sunghyo Chung$^2$ 
~~Seungjoo Yoo$^1$
~~Jaegul Choo$^1$\\
$^1$Korea University
~~$^2$Kakao Corp.}

\vspace{-1.5em}

\twocolumn[{%
\renewcommand\twocolumn[1][]{#1}%
\maketitle
\begin{center}
\includegraphics[width=0.95\linewidth]{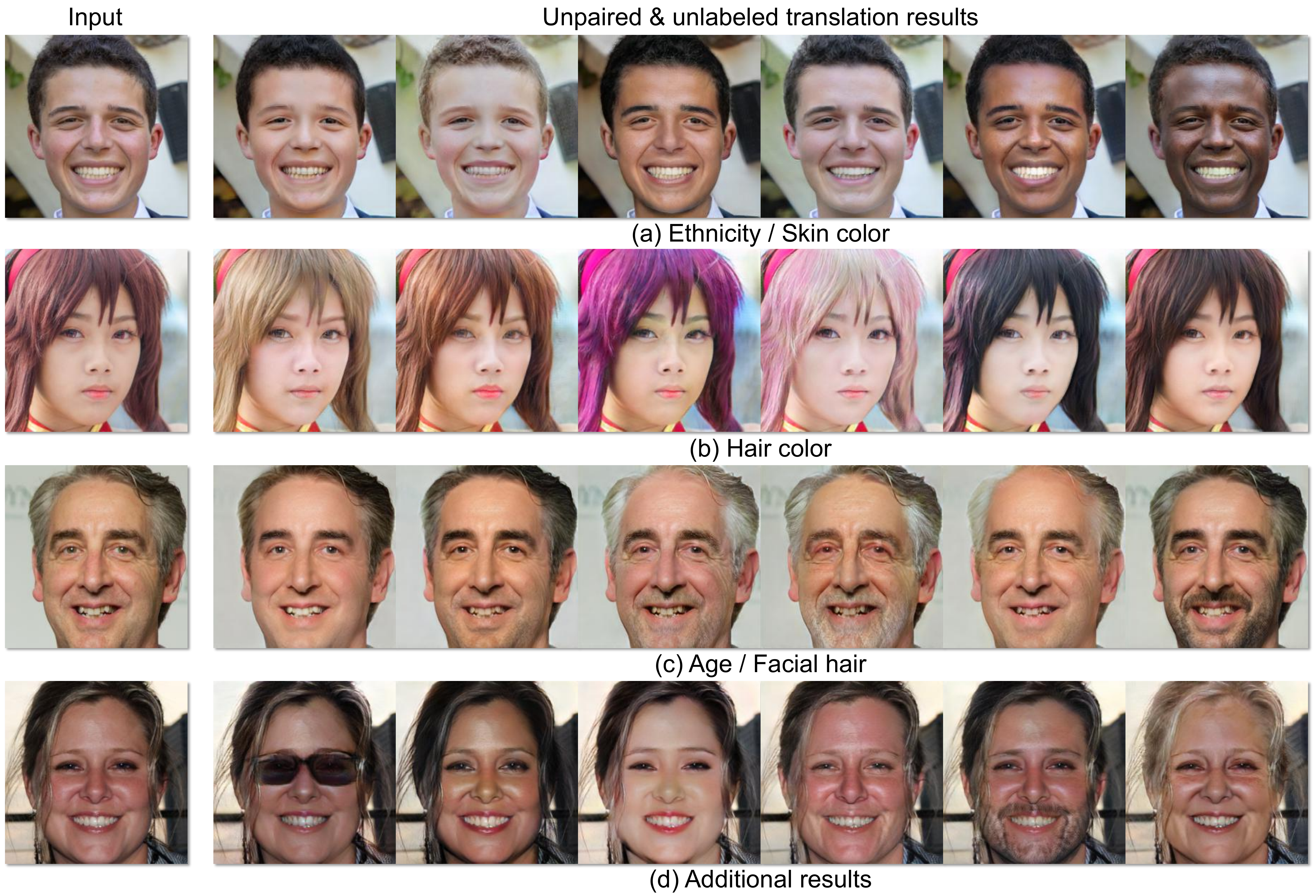}
\captionof{figure}{
Given raw, unlabeled data, our algorithm discovers novel facial attributes and performs high-quality multi-domain image translation. All results are based on \textit{newly-found} attributes from our algorithm (e.g., a wide rage of ethnicity, skin and hair color, age, facial hair, accessories, makeup). We did not use any pre-defined attribute labels to generate the results.}
\vspace{+1.0em}
\label{fig:teaser}
\end{center}
}]

\begin{abstract}
   \vspace{-1.0em}
Despite remarkable success in unpaired image-to-image translation, existing systems still require a large amount of labeled images. This is a bottleneck for their real-world applications; in practice, a model trained on labeled CelebA dataset does not work well for test images from a different distribution -- greatly limiting their application to unlabeled images of a much larger quantity. In this paper, we attempt to alleviate this necessity for labeled data in the facial image translation domain. We aim to explore the degree to which you can discover novel attributes from unlabeled faces and perform high-quality translation. To this end, we use prior knowledge about the visual world as guidance to discover novel attributes and transfer them via a novel normalization method. Experiments show that our method trained on unlabeled data produces high-quality translations, preserves identity, and be perceptually realistic as good as, or better than, state-of-the-art methods trained on labeled data.
\end{abstract}

\section{Introduction}

In recent years, unsupervised image-to-image translation has improved dramatically~\cite{CycleGAN2017, StarGAN2018, DRIT, huang2018munit}. Existing translation methods use the term \textit{unsupervised} for translating with \textit{unpaired} training data (i.e., provided with images in domain X and Y, with no information on which $x$ matches which $y$). However, existing systems, in essence, are still trained with supervision, as they require large amount of \textit{labeled} images to perform translation. This becomes a bottleneck for their application in the real world; in practice, a model trained on labeled CelebA dataset~\cite{liu2015faceattributes} does not work well for images of different test distribution due to dataset bias~\cite{torralba2011unbiased, wang2019detecting}. For instance, a model trained on CelebA images are biased towards Western, celebrity faces, which necessitates collecting, labeling, and training with new data to match a different test distribution. Hence, need for labels greatly limits their application to unlabeled images of much larger quantity.

In this paper, we attempt to alleviate the necessity for labeled data by automatically discovering novel attributes from unlabeled images -- moving towards \textit{unpaired} and \textit{unlabeled} multi-domain image-to-image translation. In particular, we focus on image translation of facial images, as they require annotation of multiple attributes (e.g., 40 attributes for 202,599 images in CelebA), which makes labeling labor- and time-intensive. While existing benchmark datasets attempt to label as much attributes as they can, we notice that much is still unnamed, e.g., CelebA only contains `pale skin' attribute among all possible skin colors. This makes us wonder: \textit{can't we make the attributes ``emerge'' from data?}

This paper aims to explore the degree to which you can discover novel attributes from unlabeled faces $X$, thus proposing our model XploreGAN. To this end, we utilize pre-trained CNN features -- making the most out of \textit{what we have already learned} about the visual world. Note that classes used for CNN pre-training (ImageNet classes) differ from the unlabeled data (facial attributes). The goal is to transfer not its specific classes, but the general knowledge on what properties makes a good class in general~\cite{han2019learning}. We use it as guidance to group new set of unlabeled faces, where each group contains a common attribute, and transfer that attribute to an input image by our newly proposed attribute summary instance normalization (ASIN). Unlike previous style normalization methods that generate affine parameters from a single image~\cite{dumoulin2017learned, huang2017arbitrary}, resulting in translation of \textit{entangled} attributes (i.e., hair color, skin color, and gender) that exist in the style image, ASIN summarizes the common feature (i.e., blond hair) among a group of images (cluster) and only transfers its common attribute (style) to the input (content). Experiments show that XploreGAN trained on \textit{unlabeled} data produces high-quality translation results as good as, or better than, state-of-the-art methods trained with \textit{labeled} data. To the best of our knowledge, this is the first method that moves towards both \textit{unpaired} and \textit{unlabeled} image-to-image translation.



\section{Proposed Method}
While existing methods uses facial images that annotates a single image with multiple labels (i.e., one-to-many mapping) to achieve multi-domain translation, we slightly modify this assumption to achieve a high-quality performance without utilizing any attribute labels. We first utilize pre-trained feature space as guidance to cluster unlabeled images by their common attribute. Using the cluster assignment as \textit{pseudo-label}, we utilize our newly proposed attribute summary instance normalization (ASIN) to summarize the common attribute (e.g., blond hair) among images in each cluster and perform high-quality translation.

\subsection{Clustering for attribute discovery}
CNN features pre-trained on ImageNet~\cite{deng2009imagenet} have been used to assess \textit{perceptual similarity} among images~\cite{johnson2016perceptual, zhang2018unreasonable}. In other words, images with similar pre-trained features are perceived as similar to humans. Exploiting this property, we propose to discover novel attributes that exist in unlabeled data by clustering their feature vectors obtained from pre-trained networks and using these cluster assignments as our \textit{pseudo-label} for attributes. In other words, we utilize the pre-trained feature space as guidance to group images by their dominant attributes. 

We adopt a standard clustering algorithm, \textit{k}-means, and partition the features from pre-trained networks $\{f(x_{1}),...,f(x_{n})\}$ into \textit{k} groups by solving:
\begin{equation}
\begin{aligned}
\min_{\mu}\min_{C}\mathop{\sum_{i=1}^{k}\sum_{x\in C_{i}}\left\Vert f(x)-\mu_{i}\right\Vert _{2}^{2}}.
\end{aligned}
\end{equation}
Solving this problem results in a set of cluster assignments $C$, centroids $\mu$, and their standard deviations $\sigma$. We use the $C$ as pseudo-labels for training the auxiliary classifier of the discriminator and use $\mu$ and $\sigma$ for conditioning the normalization layer of the generator.

\subsection{Attribute summary instance normalization}

Normalization layers play a significant role in modeling style. As~\cite{huang2017arbitrary} puts it, a single network can ``generate images in completely different styles by using the \textit{same} convolutional parameters but \textit{different} affine parameters in instance normalization (IN) layers". In other words, to inject a style to a content image, it is sufficient to simply tune the scaling and shifting parameters specific to each style after normalizing the content image. 

Previous style normalization methods generate affine parameters from a single image instance~\cite{dumoulin2017learned, huang2017arbitrary}, resulting in translation of \textit{entangled} attributes (e.g., hair color/shape, skin color, and gender) that exist in the given style image. In contrast, our approach summarizes and transfers the \textit{common attribute} (e.g., blond hair) within a group of images by generating affine parameters from the feature statistics of each cluster. We call this \textit{attribute summary instance normalization} (ASIN). We use a multilayer perceptron (MLP) $f$ to map cluster statistics to the affine parameters of the normalization layer, defined as
\begin{equation}
\begin{aligned}
\mathrm{ASIN}(x;\mu_{k},\sigma_{k})=\mathrm{\mathit{f}_{\sigma}}(\sigma_{k})\left(\frac{x-\mu(x)}{\sigma(x)}\right)+\mathrm{\mathit{f}_{\mu}}(\mu_{k}).
\end{aligned}
\end{equation}

As the generator is trained to generalize the common feature among each subset of images (cluster), ASIN allows us to discover multiple attributes in unlabeled data. ASIN can also be used in \textit{supervised} settings to summarize the common attribute among images with the same label (e.g., black hair). You may generate affine parameters from both the centroid and variance of each cluster, only the centroid information, or the domain pseudo-label (i.e., cluster assignments). We will use the first option in subsequent equations in the paper, so as not to confuse the readers.

\subsection{Objective function}
\paragraph{Cluster classification loss.}
To translate an input image $x$ to a target domain $k$, we adopt a domain classification loss~\cite{StarGAN2018} to generate images that are properly classified as its target domain. However, we use cluster assignments as \textit{pseudo-labels} for each attribute unlike previous multi-domain translation approaches that utilize pre-given labels for classification~\cite{StarGAN2018, pumarola2018ganimation}. We optimize the discriminator $D$ to classify real images $x$ to its original domain $k'$ by the loss function defined as
\begin{equation}
\begin{aligned}
\mathcal{L_{\mathit{cls}}^{\mathrm{\mathit{r}}}}=\mathbb{E_{\mathit{x,k'}}\mathit{\mathrm{\left[-\mathrm{log\mathit{D_{cls}\mathrm{(\mathit{k'\mid x})}}}\right]}}}.
\end{aligned}
\end{equation}
Similarly, we optimize the generator $G$ to classify fake images $G(x,\mu_{k},\sigma_{k})$ to its target domain $k$ via the loss function defined as
\begin{equation}
\begin{aligned}
\mathcal{L_{\mathit{cls}}^{\mathrm{\mathit{f}}}}=\mathbb{E_{\mathit{x,k}}\mathit{\mathrm{\left[-\mathrm{log\mathit{D_{cls}\mathrm{(\mathit{k\mid G(x,\mu_{k},\sigma_{k}})}}}\right]}}}.
\end{aligned}
\end{equation}
The cluster statistics act as conditional information for translating images to its corresponding pseudo domain.

\paragraph{Reconstruction and latent loss.} Our generator should be sensitive to change in content but robust to other variations. To make translated images preserve the content of its input images while changing only the domain-related details, we adopt a cycle consistency loss~\cite{kim2017learning, CycleGAN2017} to the generator, defined as 
\begin{equation}
\begin{aligned}
\mathcal{L_{\mathit{rec}}}=\mathbb{E_{\mathrm{\mathit{x,k,k'}}}}\left[\left\Vert x-G(G(x,\mu_{k},\sigma_{k}),\mu_{k'},\sigma_{k'})\right\Vert _{1}\right],
\end{aligned}
\end{equation}
where the generator is given the fake image $G(x,\mu_{k},\sigma_{k})$ and the original cluster statistics $\mu_{k'},\sigma_{k'}$ and aims to reconstruct the original real image $x$. We use the L1 norm for the reconstruction loss. 

However, solely using the pixel-level reconstruction loss does not guarantee that translated images preserve the high-level content of its original images in settings where a single generator has to learn a large number of domains simultaneously (e.g., more than 40). Inspired by~\cite{yang2018learning}, we adopt the latent loss, where we minimize the distance between real and fake images in the feature space:
\begin{equation}
\begin{aligned}
\mathcal{L_{\mathit{lnt}}}=\mathbb{E_{\mathrm{\mathit{x,k,k'}}}}\left[\left\Vert h(x)-h(G(x,\mu_{k},\sigma_{k}))\right\Vert _{2}\right].
\end{aligned}
\end{equation}
 We denote $h$ as the encoder of $G$ and use the $L_2$ norm for the latent loss. The latent loss ensures that the real and fake images have similar high-level feature representations (i.e., perceptually similar) even though they may be quite different at a pixel level.

\paragraph{Adversarial loss.} We adopt the adversarial loss used in GANs to make the generated images indistinguishable from real images. The generator $G$ tries to generate a realistic image $G(x,\mu_{k},\sigma_{k})$ given the input image $x$ and the target cluster statistics $\mu_{k},\sigma_{k}$, while the discriminator $D$ tries to distinguish between generated images and real images. To stabilize GAN training, we adopt the Wasserstein GAN objective with gradient penalty~\cite{arjovsky2017wasserstein, gulrajani2017improved}, defined as 
\begin{equation}
\begin{aligned}
\mathcal{L_{\mathit{adv}}}=\mathbb{E_{\mathrm{\mathit{x}}}}\left[D_{adv}(x)\right]-\mathbb{E_{\mathrm{\mathit{x},k}}}\left[D_{adv}(G(x,\mu_{k},\sigma_{k}))\right] \\
-\lambda_{gp}\mathbb{E_{\mathit{\hat{x}}}}\left[\left(\left\Vert \nabla_{\hat{x}}D_{adv}(\hat{x})\right\Vert _{2}-1\right)^{2}\right],
\end{aligned}
\end{equation}
where $\hat{x}$ is sampled uniformly from straight lines between pairs of real and fake images. 

\paragraph{Full objective function.} Finally, our full objective function for $D$ and $G$ can be written as
\begin{equation}
\begin{aligned}
\mathcal{L_{\mathit{D}}}=-\mathcal{L_{\mathit{adv}}}+\lambda_{cls}\mathcal{L_{\mathit{cls}}},
\end{aligned}
\end{equation}
\begin{equation}
\begin{aligned}
\mathcal{L_{\mathit{G}}}=\mathcal{L_{\mathit{adv}}}+\lambda_{cls}\mathcal{L_{\mathit{cls}}}+\lambda_{rec}\mathcal{L_{\mathit{rec}}}+\lambda_{lnt}\mathcal{L_{\mathit{lnt}}}.
\end{aligned}
\end{equation}
The hyperparameters control the relative importance of each loss function. In all experiments, we used $\lambda_{gp}=10$, $\lambda_{rec}=10$, and $\lambda_{lnt}=10$. At test time, we use the pseudo-labels to generate translated results. We are surprised to find that the pseudo-labels correspond to meaningful facial attributes; results are demonstrated in Section~\ref{sec:experiments}. 

\subsection{Implementation details}
\paragraph{Clustering stage.}  
We use the final convolutional activations (i.e., \texttt{conv5} for BagNet-17 and ResNet-50) to cluster images according to high-level attributes. We use BagNet-17~\cite{brendel2018approximating} pre-trained on ImageNet (IN)~\cite{deng2009imagenet} as the feature extractor for FFHQ~\cite{karras2018style} and CelebA~\cite{liu2015faceattributes} dataset, and ResNet-50~\cite{he2016deep} pre-trained on Stylized ImageNet (SIN)~\cite{geirhos2018imagenet} as the feature extractor of EmotioNet~\cite{fabian2016emotionet} dataset. The former is effective in detecting local texture cues, while the latter ignores texture cues but detects global shapes effectively. For clustering, the extracted features are $L_{2}$-normalized and PCA-reduced to 256 dimensions. We utilize the $k$-means implementation by Johnson et al.~\cite{JDH17}, with $k=50$ for images with $256\times 256$ resolution and $k=100$ for images with $128 \times 128$ resolution. 

\paragraph{Translation stage.}  Adapted from StarGAN~\cite{StarGAN2018}, our encoder has two convolutional layers for downsampling followed by six residual blocks~\cite{he2016deep} with spectral normalization~\cite{miyato2018spectral}. Our decoder has six residual blocks with attribute summary instance normalization (ASIN), with per-pixel noise~\cite{karras2018style} added after each convolutional layer. It is followed by two transposed convolutional layers for upsampling. We also adopt stochastic variation~\cite{karras2018style} to increase generation performance on fine, stochastic details of the image. For the discriminator, we use PatchGANs~\cite{li2016precomputed, isola2017image, zhu2017unpaired} to classify whether image patches are real or fake. As a module to predict the affine parameters for ASIN, our multi-layer perceptron consists of seven layers for FFHQ and EmotioNet datasets and three layers for the CelebA dataset. For training, we use the Adam optimizer, a mini-batch size of 32, a learning rate of 0.0001, and decay rates of $\beta_{1}=0.5$, $\beta_{2}=0.999$.

\begin{figure}[t]
\begin{center}
\includegraphics[width=1\linewidth]{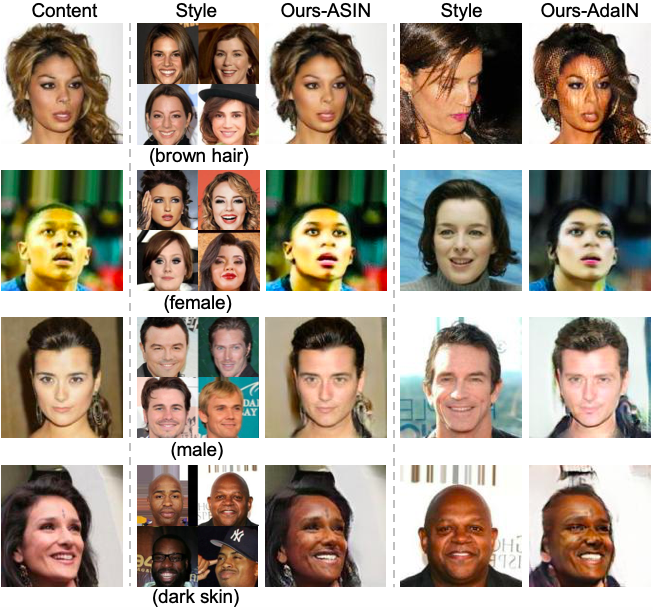}
\end{center}
\vspace{-1.5em}
  \caption{\textbf{Comparison on style normalization methods.} As AdaIN is conditioned on a single image instance to transfer style, it tends translate entangled attributes of the style image (last three rows). In contrast, ASIN summarizes a common attribute within a group (cluster) of images and transfers its specific feature, while keeping all other attributes (identity) of the content image intact.}
\label{fig:gin_vs_adain}
\end{figure}

\begin{figure*}[t]
\begin{center}
\includegraphics[width=1\linewidth]{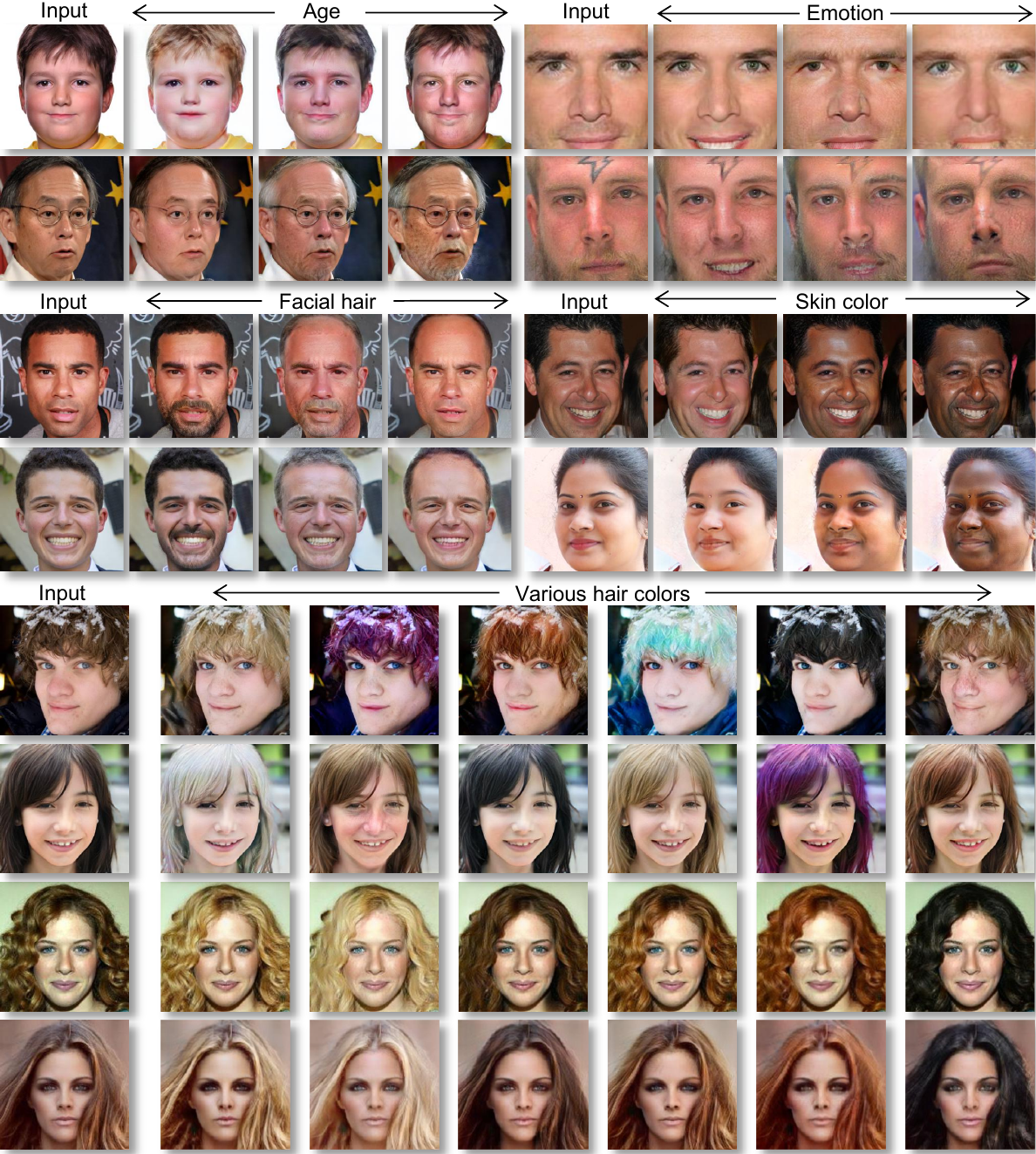}
\end{center}
  \caption{\textbf{Translation results from multiple datasets.} XploreGAN can discover various attributes in data such as diverse hair colors, ethnicity, degree of age, and facial expressions from unlabeled images. Note that labels in the figure are assigned post-hoc to enhance the interpretability of the results.}
\label{fig:novel_domains}
\end{figure*}

\begin{figure*}[t]
\begin{center}
\includegraphics[width=1\linewidth]{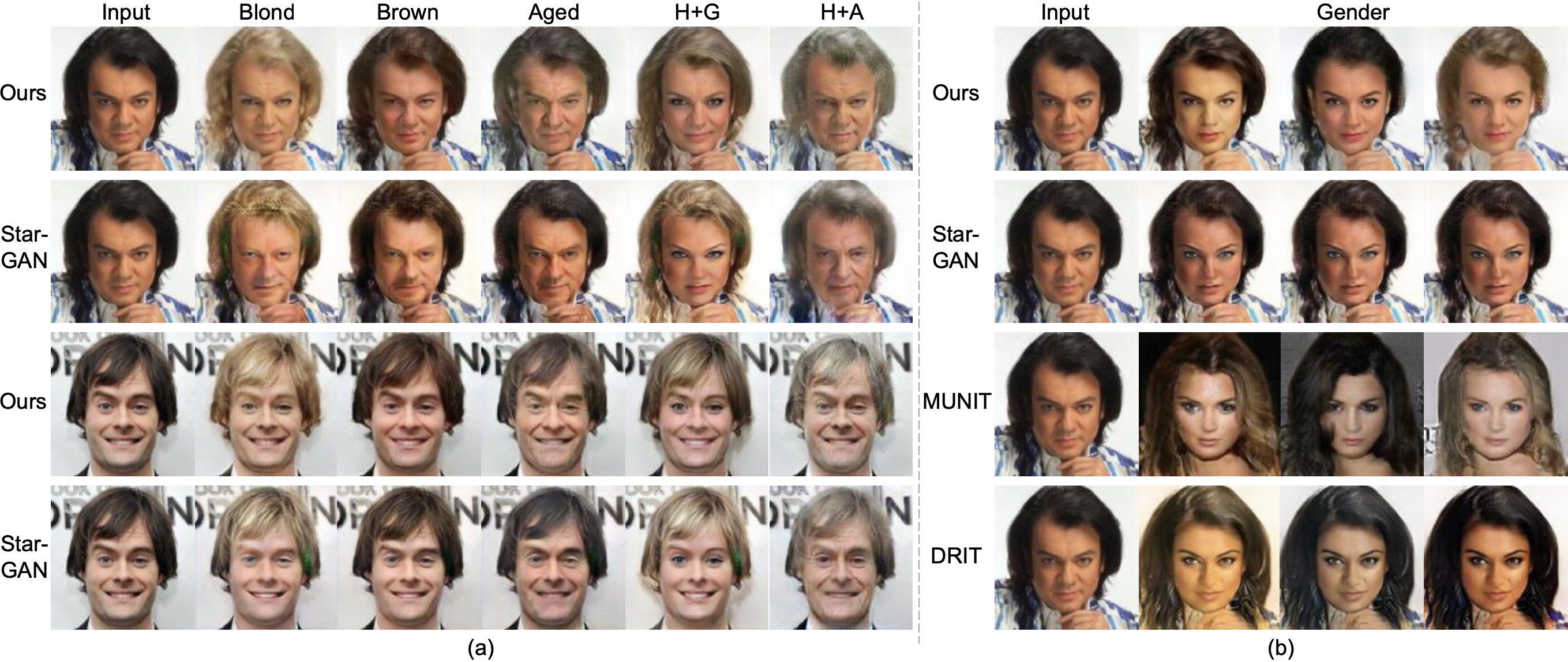}
\end{center}
  \caption{\textbf{Baseline comparisons.} Facial attribute translation results on the CelebA dataset. (a) compares multi-domain translation quality (H: hair color, G: gender, A: aged), and (b) compares multimodal tranlation quality. Each result of our model is generated from single cluster statistics.}
\label{fig:baseline_comparison}
\end{figure*}

\section{Experiments}
\label{sec:experiments}

\subsection{Datasets}
\noindent
\textbf{Flickr-Faces-HQ (FFHQ)~\cite{karras2018style}} is a high-quality human face image dataset with 70,000 images, offering a large variety in age, ethnicity, and background. The dataset is not provided with any attribute labels.
\vspace{2mm}

\noindent
\textbf{CelebFaces Attributes (CelebA)}~\cite{liu2015faceattributes} is a large-scale face dataset with 202,599 celebrity images, each annotated with 40 binary attribute labels. In our experiments, we do not utilize the attribute labels for training our model.
\vspace{2mm}

\noindent
\textbf{EmotioNet~\cite{fabian2016emotionet}} contains 950,000 face images with diverse facial expressions. The facial expressions are annotated with action units, yet we do not utilize them for training our model.

\subsection{Baseline models}
We compare with baseline models that utilize \textit{unpaired} yet \textit{labeled} datasets. All experiments of the baselines are conducted using the original codes and hyperparameters. As XploreGAN does not use any labels during training, at test time, we select pseudo-labels that best estimates the labels used by other baseline models (e.g., pseudo-label that best corresponds to `blond'). Each result of our model is generated from a single cluster statistics. 

\vspace{2mm}
\noindent
\textbf{StarGAN} is a state-of-the-art \emph{multi-domain} image translation model that uses the attribute label during training.

\vspace{2mm}
\noindent
\textbf{DRIT and MUNIT} are state-of-the-art models that perform \emph{multimodal} image translation between two domains.

\subsection{Comparison on style normalization} 
We show qualitative comparison of group-based ASIN and instance-based AdaIN. For fair comparison, we substitute the ASIN layers with AdaIN as implemented in \cite{huang2018munit} while maintaining all the other network architecture and training settings. As shown in Fig.~\ref{fig:gin_vs_adain}, AdaIN depends on a single image instance to transfer style. AdaIN results in translation of entangled attributes (e.g., hair color/shape, gender, background color; last three rows of Fig.~\ref{fig:gin_vs_adain}) that exist within the reference image. In contrast, ASIN is able to summarize the common attribute within a group of images (e.g., hair color) and transfer its specific attribute. This makes it easy for users to transfer a particular attribute they desire while preserving all other attributes (identity) of the content image intact.

\subsection{Qualitative evaluation}
As shown in Fig.~\ref{fig:baseline_comparison}, we qualitatively compare face attribute translation results on the CelebA dataset. All baseline models are trained using the attribute labels, while XploreGAN is trained with unlabeled data. As we increase the number of $k$, we can discover multiple subsets of a single attribute (e.g., diverse styles of `women'; further discussed in Section~\ref{clus_experiments}). This can be thought as discovering $k$ modes in data. Thus, we can compare our model to not only multi-domain translation but also multimodal translation between two domains. Fig.~\ref{fig:baseline_comparison} shows that our method can enerate translation results as high quality as other models trained with labels. Also, Fig.~\ref{fig:novel_domains} shows that XploreGAN can perform high-quality translation for various datasets (FFHQ~\cite{karras2018style},  CelebA~\cite{liu2015faceattributes} and Emotionet~\cite{fabian2016emotionet}). We present additional qualitative results in the Appendix.

\subsection{Quantitative evaluation}
A high-quality image translation should i) transfer the target attribute well while ii) preserving identity of the input image and iii) look realistic to human eyes. We quantitatively measure the three quality metrics by attribute classification, face verification, and a user study. 

\vspace{3mm}
\noindent
\textbf{Attribute classification.}
To measure how well a model transfers attributes, we compare classification accuracy of synthesized images on face attributes. We train a binary classifier for each of the selected attributes (blond, brown, old, male, and female) in the CelebA dataset (70\%/30\% split for training and test set), which results in an average accuracy of 95.8\% on real test images. We train all models with the same training set and perform image translation on the same test set. Finally, we measured classification accuracy of translated images using the trained classifier above. Surprisingly, XploreGAN outperforms all baseline models in almost all attribute translation as shown in Table~\ref{table:cls_results}. This shows that our method trained on \textit{unlabeled} data can perform high-quality translation as good as, or sometimes even better than, those models trained on \textit{labeled} data.

\begin{table}[]
\begin{tabular}{lccccc}
\hline
Method     & Blond  & Brown  & Aged  & Male & Female \\ \hline
Ours       & \textbf{90.2}   & 77.4   & \textbf{90.0} & \textbf{99.7} & \textbf{99.6}   \\
StarGAN    & 90.0   & \textbf{86.1}   & 88.4 & 97.5 & 98.0   \\
MUNIT      & -      & -      & -    & 95.7 & 99.1   \\
DRIT       & -      & -      & -    & 98.8 & 98.5   \\ \hline
Real Image & 97.2   & 92.4   & 93.3 & 98.5 & 97.6   \\ 
\hline
\end{tabular}
\vspace{0.1em}
\caption{Classification performance for translated images, evaluated on five CelebA attributes.}
\label{table:cls_results}
\end{table}

\vspace{3mm}
\noindent
\textbf{Identity preservation.}
We measure the identity preservation performance of translated images using a state-of-the-art face verification model. We use ArcFace~\cite{deng2018arcface} pre-trained on Celeb-1M dataset~\cite{guo2016msceleb}, which shows an average accuracy of 89.76\% on the CelebA test set. Then, we perform image translation on the same unseen test set regarding five face attributes (blond hair, brown hair, aged, male, and female). To measure how well a translated image preserves identity of the input image, we measure face verification accuracy on pairs of real and fake images using the pre-trained verification model above. As shown in Table~\ref{table:id_results}, our method produces translation results that preserve identity of the input image as good as, or sometimes even better than, most baseline models trained on attribute labels. Although multimodal image translation models (MUNIT and DRIT) show high classification performance (i.e., they transfer target attributes well), we observe that they tend modify the input to the extent that it greatly hinders identity preservation. 

\begin{table}[]
\begin{tabular}{lccccc}
\hline
Method  & Blond & Brown & Aged   & Male  & Female \\ \hline
Ours    & \textbf{99.3}  & \textbf{99.4}  & \textbf{99.1} & 90.1 & \textbf{94.8}   \\
StarGAN & 96.8 & 99.0 & 98.8 & \textbf{97.5} & 93.7      \\
MUNIT   & -     & -     & -     & 9.7     & 16.3      \\
DRIT    & -     & -     & -     & 72.2     & 62.0      \\ 
\hline
\end{tabular}
\vspace{0.1em}
\caption{Identity preservation performance of translated images from different methods shown by facial verification accuracy.}
\label{table:id_results}
\end{table}

\begin{table}[]
\centering
\begin{tabular}{lcccccc}
\hline
Method   & Hair          & Aged        & Gender           & H+G   & H+A      \\ \hline
Ours    & \textbf{54.7} & 43.8          & \textbf{64.5}    & \textbf{89.6} & \textbf{53.1} \\
StarGAN & 45.3          & \textbf{56.2} & 14.6            & 10.4          & 46.9           \\
MUNIT   & -             & -             & 4.2             & -             & -             \\
DRIT    & -             & -             & 16.7              & -             & -             \\ \hline
\end{tabular}
\vspace{0.1em}
\caption{Results from the user study. Last two columns correspond to simultaneous translations of multiple domains. (H+G: Hair+Gender, H+A: Hair+Aged)}
\label{table:user_study}
\end{table}

\vspace{3mm}
\noindent
\textbf{User study.} To evaluate whether the translated outputs look realistic to human eyes, we conduct a user study with 32 participants. Users are asked to choose which output is most successful in producing high-quality images, while preserving content and transferring the target attribute well. 20 questions were given for each of the six attributes, with a total of 120 questions. Note that MUNIT and DRIT produces multimodal outputs, thus a single image is randomly chosen for the user study. Table~\ref{table:user_study} shows that our model performs as good as supervised models across diverse attributes. Though StarGAN achieves promising results, its results on H+G frequently has green artifacts, which decreases user preference. 
\subsection{Analysis on the clustering stage} \label{clus_experiments}
\begin{figure}[t]
\subfloat[BagNet-17 (IN)]
{\includegraphics[width=0.48\linewidth]{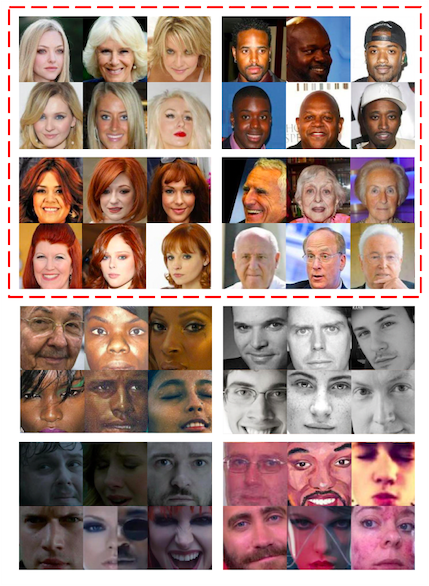}}
\hspace*{\fill} 
\subfloat[ResNet-50 (SIN)]
{\includegraphics[width=0.48\linewidth]{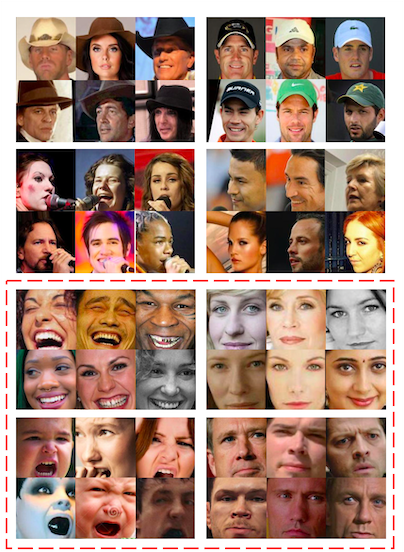}}
  \caption{\textbf{Comparing different pre-trained feature spaces.} Different pre-trained feature spaces provide highly different attribute clusters. (a) Texture-based representation: As ImageNet-pre-trained BagNets are constrained to look at small local features, it is effective in detecting minor texture cues (e.g., skin color, age, hair color, and lighting). (b) Shape-biased representation: ResNets trained on Stylized-ImageNet (SIN) are effective in ignoring texture cues and focusing on global shape information (e.g., facial expressions, gestures, and viewpoints). We used BagNet-17 as the feature extractor for the CelebA dataset (first four rows) and ResNet-50 pre-trained on SIN for the EmotioNet dataset (last four rows).}
\label{fig:feature_extractor}
\end{figure}

\noindent
\textbf{Comparison on pre-trained feature spaces.}
The pre-trained feature space provides a guideline to group the unlabeled images. 
We found that difference in model architectures and datasets it was pre-trained on leads to significantly different feature spaces, i.e., representation bias. We attempt to exploit this ``skewness” towards recognition of certain types of features (e.g., texture or shape) -- to group novel images in different directions. We will mainly compare two feature spaces: texture-biased BagNets and shape-biased ResNets. It has been found that ImageNet pre-trained CNNs are strongly biased towards recognizing textures rather than shapes~\cite{geirhos2018imagenet}. In relation to this characteristic, BagNets~\cite{brendel2018approximating} are designed to be more sensitive to recognizing local textures compared to vanilla ResNets~\cite{he2016deep} by limiting the receptive field size. They are designed to focus on small local image features without taking into account their larger spatial relationships. On the other hand, ResNets trained on Stylized ImageNet~\cite{geirhos2018imagenet} (denoted ResNet (SIN)) ignores texture cues altogether and focuses on global shapes of images. 

Fig.~\ref{fig:feature_extractor} shows the characteristics of these two feature spaces. BagNets trained on ImageNet are effective in detecting detailed texture cues (e.g., skin color and texture, degree of age, hair color/shape, lighting). However, BagNets or vanilla ResNets are ineffective in detecting facial emotions, as they tend to produce clustering results biased towards local texture cues. For this purpose, we found that ResNet (SIN) is highly effective in ignoring texture cues and focusing on global shape information (e.g., facial expressions, gestures, view-points). Considering these qualities, we have adopted BagNet-17 (IN) as the feature extractor for the CelebA~\cite{liu2015faceattributes} and FFHQ~\cite{karras2018style} dataset, and ResNet-50 (SIN) for the EmotioNet~\cite{fabian2016emotionet} dataset.

\begin{figure}[t]
\begin{center}
   \includegraphics[width=0.95\linewidth]{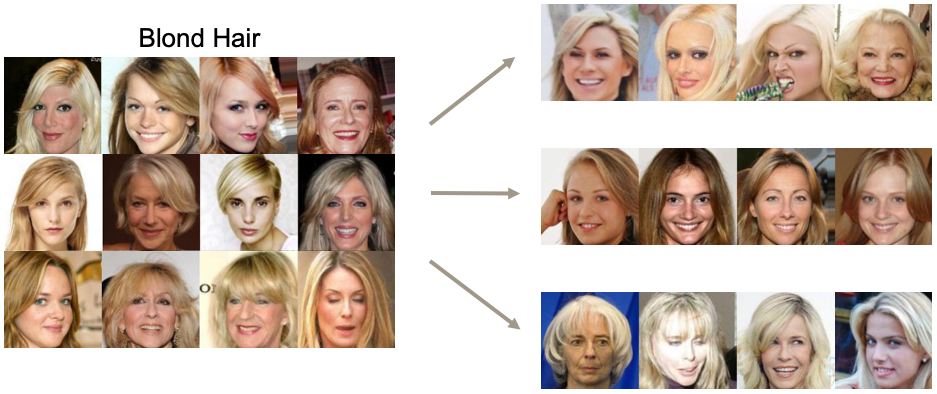}
\end{center}
\vspace{-1.5em}
   \caption{\textbf{Effects of increasing the number of clusters.} One can discover hidden attribute subsets previously entangled in a single cluster.}
\label{fig:blond_type}
\end{figure}


\vspace{3mm}
\noindent
\textbf{Choosing the number of clusters.} As shown in Fig.~\ref{fig:blond_type}, a single blond cluster is subdivided to specific types of blond hair as the number of $k$ increases. As such, small values of \textit{k} produce compact clusters with highly distinctive features, while large values of \textit{k} produce clusters with similar yet detailed features. In reality, there is no ground truth of the optimal number of $k$. In other words, how a human labeler defines a single attribute in a given dataset is highly subjective (e.g., `pale makeup' itself can be a single attribute, or it may be further subdivided into `pale skin', `wearing eyeshadow', and `wearing lipstick' depending on the labeler's preference). In our model, users can indirectly control such degree of division by adjusting the number of $k$.

\section{Related Work}
\noindent
\textbf{Generative adversarial networks (GANs).} GANs~\cite{goodfellow2014generative} have achieved remarkable success in image generation. The key to its success is the adversarial loss, where the discriminator tries to distinguish between real and fake images while the generator tries to fool the discriminator by producing realistic fake images. Several studies leverage \textit{conditional} GANs in order to generate samples conditioned on the class~\cite{mirza2014conditional, odena2016semi, odena2016conditional}, text description~\cite{reed2016generative, han2017stackgan, Tao18attngan}, domain information~\cite{StarGAN2018, pumarola2018ganimation}, input image~\cite{isola2017image}, or color features~\cite{bahng2018coloring}. In this paper, we adopt the adversarial loss conditioned on cluster statistics to generate corresponding translated images indistinguishable from real images.

\vspace{3mm}
\noindent
\textbf{Unpaired image-to-image translation.}
Image-to-image translation~\cite{isola2017image, zhu2017toward} has recently shown remarkable success. CycleGAN~\cite{CycleGAN2017} extends image-to-image translation to unpaired settings, which broadens application of deep learning models to more datasets. Multi-domain image-to-image translation models~\cite{StarGAN2018, pumarola2018ganimation} propose methods that generate diverse outputs when given domain labels. DRIT~\cite{DRIT} and MUNIT~\cite{huang2018munit} further develop image translation models to produce random multimodal outputs using unpaired data. Most existing image-to-image translation models rely on labeled data. Unlike previous approaches that define the term `unpaired' as synonymous to unsupervised, we define unsupervised to encompass both unpaired and unlabeled. According to our definition, no previous work on image-to-image translation has tackled such settings. 

\vspace{3mm}
\noindent
\textbf{Clustering for discovering the unknown.} Clustering is a powerful unsupervised learning method that groups data by their similarity. Clustering is used to discover novel object classes in images~\cite{liu2016unsupervised} and videos~\cite{osep2019large, triebel2010segmentation, herbst2011rgb, xie2018object}. Instead of discovering new object classes, our work aims to discover attributes within unlabeled data through clustering. Finding attributes is a complicated task, as a single image can have multiple different attributes. To the best of our knowledge, our work is the first to perform image-to-image translation using newly discovered attributes from unlabeled data.  

\vspace{3mm}
\noindent
\textbf{Instance normalization for style transfer.} To ease training of neural networks, batch normalization (BN) was originally introduced. BN normalizes each feature channel by the mean and standard deviation from mini-batches of images. Instance normalization~\cite{ulyanov2016instance} samples the mean and standard deviation from each sample. Extending IN, conditional instance normalization~\cite{dumoulin2017learned} learns different sets of parameters for each style. Adaptive instance normalization (AdaIN)~\cite{huang2017arbitrary} performs normalization without additional trainable parameters, to which MUNIT adds trainable parameters for stronger translation ability. In contrast to existing normalization methods that perform style transfer on image instances, our attribute summary instance normalization (ASIN) uses cluster statistics to summarize the common attribute within each cluster and allows translation of fine, detailed attributes.

\section{Conclusion}
In this paper, we attempt to alleviate the necessity for labeled data in the facial image translation domain. Provided with raw, unlabeled data, we propose an \textit{unpaired} and \textit{unlabeled} multi-domain image-to-image translation method. We utilize prior knowledge from pre-trained feature spaces to group unseen, unlabeled images. Attribute summary instance normalization (ASIN) can effectively summarize the common attribute within clusters, enabling high-quality translation of specific attributes. We demonstrate that our model can produce results as good as, or sometimes better than, most state-of-the-art methods.

{\small
\bibliographystyle{ieee_fullname}
\bibliography{egbib}

\begin{thebibliography}{10}\itemsep=-1pt

\bibitem{arjovsky2017wasserstein}
Martin Arjovsky, Soumith Chintala, and L{\'e}on Bottou.
\newblock Wasserstein generative adversarial networks.
\newblock In {\em Proceedings of the 34th International Conference on Machine
  Learning (ICML)}, 2017.

\bibitem{bahng2018coloring}
Hyojin Bahng, Seungjoo Yoo, Wonwoong Cho, David~Keetae Park, Ziming Wu,
  Xiaojuan Ma, and Jaegul Choo.
\newblock Coloring with words: Guiding image colorization through text-based
  palette generation.
\newblock In {\em Proceedings of the European Conference on Computer Vision
  (ECCV)}, pages 431--447, 2018.

\bibitem{brendel2018approximating}
Wieland Brendel and Matthias Bethge.
\newblock Approximating cnns with bag-of-local-features models works
  surprisingly well on imagenet.
\newblock 2018.

\bibitem{StarGAN2018}
Yunjey Choi, Minje Choi, Munyoung Kim, Jung-Woo Ha, Sunghun Kim, and Jaegul
  Choo.
\newblock Stargan: Unified generative adversarial networks for multi-domain
  image-to-image translation.
\newblock In {\em The IEEE Conference on Computer Vision and Pattern
  Recognition (CVPR)}, June 2018.

\bibitem{deng2009imagenet}
Jia Deng, Wei Dong, Richard Socher, Li-Jia Li, Kai Li, and Li Fei-Fei.
\newblock Imagenet: A large-scale hierarchical image database.
\newblock In {\em The IEEE Conference on Computer Vision and Pattern
  Recognition (CVPR)}, pages 248--255. Ieee, 2009.

\bibitem{deng2018arcface}
Jiankang Deng, Jia Guo, Niannan Xue, and Stefanos Zafeiriou.
\newblock Arcface: Additive angular margin loss for deep face recognition.
\newblock {\em arXiv preprint arXiv:1801.07698}, 2018.

\bibitem{dumoulin2017learned}
Vincent Dumoulin, Jonathon Shlens, and Manjunath Kudlur.
\newblock A learned representation for artistic style.
\newblock 2017.

\bibitem{fabian2016emotionet}
C Fabian Benitez-Quiroz, Ramprakash Srinivasan, and Aleix~M Martinez.
\newblock Emotionet: An accurate, real-time algorithm for the automatic
  annotation of a million facial expressions in the wild.
\newblock In {\em Proceedings of the IEEE Conference on Computer Vision and
  Pattern Recognition}, pages 5562--5570, 2016.

\bibitem{geirhos2018imagenet}
Robert Geirhos, Patricia Rubisch, Claudio Michaelis, Matthias Bethge, Felix~A
  Wichmann, and Wieland Brendel.
\newblock Imagenet-trained cnns are biased towards texture; increasing shape
  bias improves accuracy and robustness.
\newblock {\em International Conference on Learning Representations (ICLR)},
  2019.

\bibitem{goodfellow2014generative}
Ian Goodfellow, Jean Pouget-Abadie, Mehdi Mirza, Bing Xu, David Warde-Farley,
  Sherjil Ozair, Aaron Courville, and Yoshua Bengio.
\newblock Generative adversarial nets.
\newblock In {\em Advances in Neural Information Processing Systems}, pages
  2672--2680, 2014.

\bibitem{gulrajani2017improved}
Ishaan Gulrajani, Faruk Ahmed, Martin Arjovsky, Vincent Dumoulin, and Aaron~C
  Courville.
\newblock Improved training of wasserstein gans.
\newblock In {\em Advances in Neural Information Processing Systems}, pages
  5767--5777, 2017.

\bibitem{guo2016msceleb}
Yandong Guo, Lei Zhang, Yuxiao Hu, Xiaodong He, and Jianfeng Gao.
\newblock M{S}-{C}eleb-1{M}: A dataset and benchmark for large scale face
  recognition.
\newblock In {\em European Conference on Computer Vision}, 2016.

\bibitem{han2019learning}
Kai Han, Andrea Vedaldi, and Andrew Zisserman.
\newblock Learning to discover novel visual categories via deep transfer
  clustering.
\newblock In {\em Proceedings of the IEEE International Conference on Computer
  Vision}, pages 8401--8409, 2019.

\bibitem{he2016deep}
Kaiming He, Xiangyu Zhang, Shaoqing Ren, and Jian Sun.
\newblock Deep residual learning for image recognition.
\newblock In {\em Proceedings of the IEEE conference on computer vision and
  pattern recognition}, pages 770--778, 2016.

\bibitem{herbst2011rgb}
Evan Herbst, Xiaofeng Ren, and Dieter Fox.
\newblock Rgb-d object discovery via multi-scene analysis.
\newblock In {\em IEEE/RSJ International Conference on Intelligent Robots and
  Systems (IROS)}, 2011.

\bibitem{huang2017arbitrary}
Xun Huang and Serge Belongie.
\newblock Arbitrary style transfer in real-time with adaptive instance
  normalization.
\newblock In {\em Proceedings of the IEEE International Conference on Computer
  Vision}, pages 1501--1510, 2017.

\bibitem{huang2018munit}
Xun Huang, Ming-Yu Liu, Serge Belongie, and Jan Kautz.
\newblock Multimodal unsupervised image-to-image translation.
\newblock In {\em Proceedings of the European Conference on Computer Vision
  (ECCV)}, 2018.

\bibitem{isola2017image}
Phillip Isola, Jun-Yan Zhu, Tinghui Zhou, and Alexei~A Efros.
\newblock Image-to-image translation with conditional adversarial networks.
\newblock In {\em Proceedings of the IEEE conference on computer vision and
  pattern recognition}, pages 1125--1134, 2017.

\bibitem{johnson2016perceptual}
Justin Johnson, Alexandre Alahi, and Li Fei-Fei.
\newblock Perceptual losses for real-time style transfer and super-resolution.
\newblock In {\em European conference on computer vision}, pages 694--711.
  Springer, 2016.

\bibitem{JDH17}
Jeff Johnson, Matthijs Douze, and Herv{\'e} J{\'e}gou.
\newblock Billion-scale similarity search with gpus.
\newblock {\em arXiv preprint arXiv:1702.08734}, 2017.

\bibitem{karras2018style}
Tero Karras, Samuli Laine, and Timo Aila.
\newblock A style-based generator architecture for generative adversarial
  networks.
\newblock {\em arXiv preprint arXiv:1812.04948}, 2018.

\bibitem{kim2017learning}
Taeksoo Kim, Moonsu Cha, Hyunsoo Kim, Jung~Kwon Lee, and Jiwon Kim.
\newblock Learning to discover cross-domain relations with generative
  adversarial networks.
\newblock In {\em Proceedings of the 34th International Conference on Machine
  Learning (ICML)}, 2017.

\bibitem{DRIT}
Hsin-Ying Lee, Hung-Yu Tseng, Jia-Bin Huang, Maneesh~Kumar Singh, and
  Ming-Hsuan Yang.
\newblock Diverse image-to-image translation via disentangled representations.
\newblock In {\em Proceedings of the European Conference on Computer Vision
  (ECCV)}, 2018.

\bibitem{li2016precomputed}
Chuan Li and Michael Wand.
\newblock Precomputed real-time texture synthesis with markovian generative
  adversarial networks.
\newblock In {\em European Conference on Computer Vision}, pages 702--716.
  Springer, 2016.

\bibitem{liu2016unsupervised}
Liangchen Liu, Feiping Nie, Teng Zhang, Arnold Wiliem, and Brian~C Lovell.
\newblock Unsupervised automatic attribute discovery method via multi-graph
  clustering.
\newblock In {\em International Conference on Pattern Recognition (ICPR)},
  pages 1713--1718. IEEE, 2016.

\bibitem{liu2015faceattributes}
Ziwei Liu, Ping Luo, Xiaogang Wang, and Xiaoou Tang.
\newblock Deep learning face attributes in the wild.
\newblock In {\em Proceedings of International Conference on Computer Vision
  (ICCV)}, 2015.

\bibitem{mirza2014conditional}
Mehdi Mirza and Simon Osindero.
\newblock Conditional generative adversarial nets.
\newblock {\em arXiv preprint arXiv:1411.1784}, 2014.

\bibitem{miyato2018spectral}
Takeru Miyato, Toshiki Kataoka, Masanori Koyama, and Yuichi Yoshida.
\newblock Spectral normalization for generative adversarial networks.
\newblock {\em ICLR}, 2018.

\bibitem{odena2016semi}
Augustus Odena.
\newblock Semi-supervised learning with generative adversarial networks.
\newblock {\em arXiv preprint arXiv:1606.01583}, 2016.

\bibitem{odena2016conditional}
Augustus Odena, Christopher Olah, and Jonathon Shlens.
\newblock Conditional image synthesis with auxiliary classifier gans.
\newblock {\em arXiv preprint arXiv:1610.09585}, 2016.

\bibitem{osep2019large}
Aljosa Osep, Paul Voigtlaender, Jonathon Luiten, Stefan Breuers, and Bastian
  Leibe.
\newblock Large-scale object mining for object discovery from unlabeled video.
\newblock {\em arXiv preprint arXiv:1903.00362}, 2019.

\bibitem{pumarola2018ganimation}
A. Pumarola, A. Agudo, A.M. Martinez, A. Sanfeliu, and F. Moreno-Noguer.
\newblock Ganimation: Anatomically-aware facial animation from a single image.
\newblock In {\em Proceedings of the European Conference on Computer Vision
  (ECCV)}, 2018.

\bibitem{reed2016generative}
Scott Reed, Zeynep Akata, Xinchen Yan, Lajanugen Logeswaran, Bernt Schiele, and
  Honglak Lee.
\newblock Generative adversarial text-to-image synthesis.
\newblock In {\em Proceedings of The 33rd International Conference on Machine
  Learning (ICML)}, 2016.

\bibitem{Tao18attngan}
Qiuyuan Huang Han Zhang Zhe Gan Xiaolei Huang Xiaodong~He Tao~Xu,
  Pengchuan~Zhang.
\newblock Attngan: Fine-grained text to image generation with attentional
  generative adversarial networks.
\newblock 2018.

\bibitem{torralba2011unbiased}
Antonio Torralba, Alexei~A Efros, et~al.
\newblock Unbiased look at dataset bias.
\newblock In {\em CVPR}. Citeseer, 2011.

\bibitem{triebel2010segmentation}
Rudolph Triebel, Jiwon Shin, and Roland Siegwart.
\newblock Segmentation and unsupervised part-based discovery of repetitive
  objects.
\newblock {\em Robotics: Science and Systems VI}, pages 1--8, 2010.

\bibitem{ulyanov2016instance}
Dmitry Ulyanov, Andrea Vedaldi, and Victor Lempitsky.
\newblock Instance normalization: The missing ingredient for fast stylization.
\newblock {\em arXiv preprint arXiv:1607.08022}, 2016.

\bibitem{wang2019detecting}
Sheng-Yu Wang, Oliver Wang, Andrew Owens, Richard Zhang, and Alexei~A Efros.
\newblock Detecting photoshopped faces by scripting photoshop.
\newblock {\em ICCV}, 2019.

\bibitem{xie2018object}
Christopher Xie, Yu Xiang, Dieter Fox, and Zaid Harchaoui.
\newblock Object discovery in videos as foreground motion clustering.
\newblock {\em arXiv preprint arXiv:1812.02772}, 2018.

\bibitem{yang2018learning}
Hongyu Yang, Di Huang, Yunhong Wang, and Anil~K Jain.
\newblock Learning face age progression: A pyramid architecture of gans.
\newblock In {\em Proceedings of the IEEE Conference on Computer Vision and
  Pattern Recognition}, pages 31--39, 2018.

\bibitem{han2017stackgan}
Han Zhang, Tao Xu, Hongsheng Li, Shaoting Zhang, Xiaogang Wang, Xiaolei Huang,
  and Dimitris Metaxas.
\newblock Stackgan: Text to photo-realistic image synthesis with stacked
  generative adversarial networks.
\newblock In {\em Proceedings of the IEEE International Conference on Computer
  Vision (ICCV)}, 2017.

\bibitem{zhang2018unreasonable}
Richard Zhang, Phillip Isola, Alexei~A Efros, Eli Shechtman, and Oliver Wang.
\newblock The unreasonable effectiveness of deep features as a perceptual
  metric.
\newblock In {\em Proceedings of the IEEE Conference on Computer Vision and
  Pattern Recognition}, pages 586--595, 2018.

\bibitem{zhu2017unpaired}
Jun-Yan Zhu, Taesung Park, Phillip Isola, and Alexei~A Efros.
\newblock Unpaired image-to-image translation using cycle-consistent
  adversarial networks.
\newblock In {\em Proceedings of the IEEE International Conference on Computer
  Vision}, pages 2223--2232, 2017.

\bibitem{CycleGAN2017}
Jun-Yan Zhu, Taesung Park, Phillip Isola, and Alexei~A Efros.
\newblock Unpaired image-to-image translation using cycle-consistent
  adversarial networkss.
\newblock In {\em Proceedings of the IEEE International Conference on Computer
  Vision (ICCV)}, 2017.

\bibitem{zhu2017toward}
Jun-Yan Zhu, Richard Zhang, Deepak Pathak, Trevor Darrell, Alexei~A Efros,
  Oliver Wang, and Eli Shechtman.
\newblock Toward multimodal image-to-image translation.
\newblock In {\em Advances in Neural Information Processing Systems}, pages
  465--476, 2017.

\end{thebibliography}
}

\end{document}